\documentclass[conference]{IEEEtran}
\usepackage[T1]{fontenc}

\usepackage{multirow}
\ifCLASSINFOpdf
   \usepackage[pdftex]{graphicx}
   
   \graphicspath{{./pdf/}{./jpeg/}{./Figure/}}
   \DeclareGraphicsExtensions{.pdf,.jpeg,.png}
\else
   \usepackage[dvips]{graphicx}
   \graphicspath{{./eps/}}
   \DeclareGraphicsExtensions{.eps}
\fi
\usepackage{mathtools}
\usepackage{color}
\usepackage{algorithm}
\usepackage{algpseudocode}

\usepackage{times}
\usepackage{epsfig}
\usepackage{graphicx}
\usepackage{amsmath}
\usepackage{amssymb}
\usepackage{romannum}
\usepackage{color}
\usepackage{caption}
\usepackage{subcaption}
\usepackage{tikz}
\usepackage{pgfplots}
\usepackage{xcolor}
\pgfplotsset{compat=1.18}

\usepackage{algorithm,algpseudocode}

\usepackage{url}

\usepackage{booktabs}
\usepackage{fontawesome5}

\begin{document}
%
\title{UA-ChatDev: Uncertainty-Aware Multi-Agent Collaboration for Reliable Software Development }




\author{\IEEEauthorblockN{Temitayo Olamilekan Ogunsusi,  Lijun Qian, and Xishuang Dong} \\
\IEEEauthorblockA{Department of Electrical and Computer Engineering \\ 
Prairie View A\&M University, 
Prairie View, TX 77446, USA \\
Email: togunsusi@pvamu.edu,  liqian@pvamu.edu, xidong@pvamu.edu}}


\newcommand{\uachatdev}{\textsc{UA-ChatDev}}
\newcommand{\chatdev}{\textsc{ChatDev}}
\newcommand{\uala}{\textsc{UALA}}

\maketitle

\begin{abstract}

Software development is a complex task that demands cooperation among agents with diverse roles. Large language models (LLMs) have enabled autonomous multi-agent software development frameworks that leverage role-based collaboration to automate requirements analysis, coding, testing, and refinement. However, existing approaches typically assume that intermediate agent outputs are equally reliable, leaving them vulnerable to hallucination propagation, where incorrect decisions generated in early development phases are transferred to downstream agents and negatively impact final software quality. To address this challenge, we propose UA-ChatDev, an uncertainty-aware multi-agent software development framework that integrates uncertainty quantification into agent interactions. It introduces a lightweight uncertainty estimation mechanism based on token-level log probabilities to assess the confidence of agent responses and employs phase-aware threshold calibration to selectively trigger retrieval-based verification when uncertainty exceeds acceptable levels. Extensive experiments on the SRDD benchmark demonstrate that UA-ChatDev consistently outperforms existing single-agent and multi-agent software development frameworks across completeness, executability, consistency, and overall quality metrics. Further ablation studies and communication analyses verify that uncertainty-aware interactions enhance code execution reliability.
\end{abstract}

\begin{IEEEkeywords} Agentic AI, Large Language Models, Software Development. Uncertainty Measurement; \end {IEEEkeywords}

%
\IEEEpeerreviewmaketitle

\section{Introduction}
\label{sec1:introduction}

\begin{figure*}[t]
    \centering
    \includegraphics[width=\textwidth]{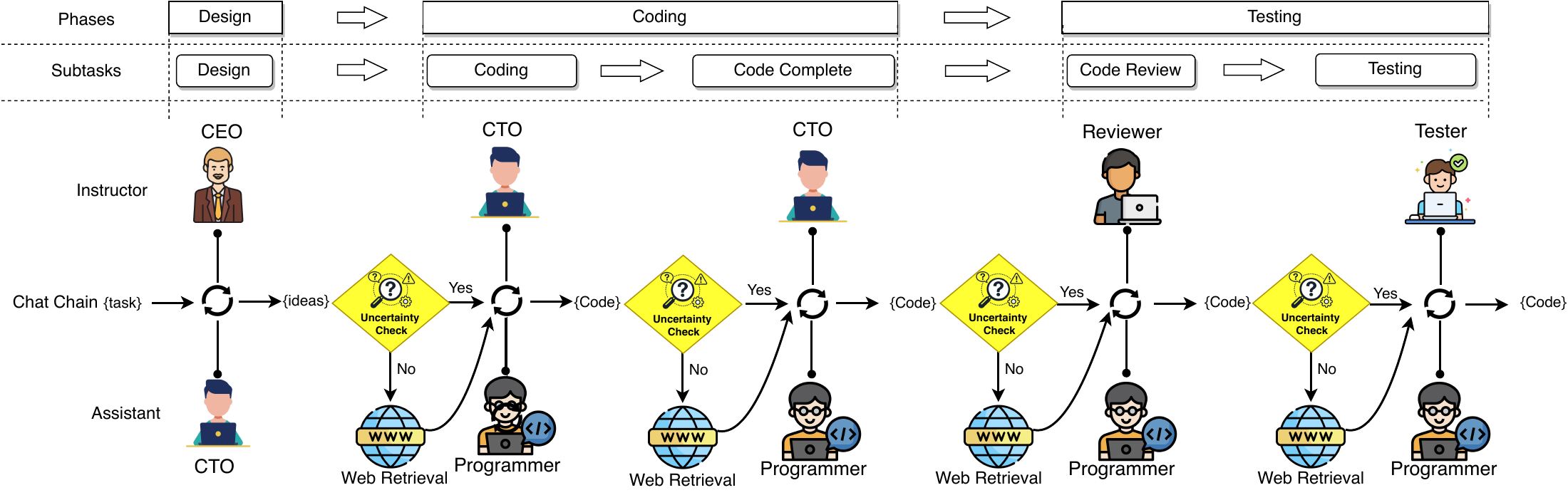}
    \caption{Illustration of the proposed method workflow that integrates ChatDev with uncertainty quantification. Given an initial task requirement, software agents collaborate through multi-turn communication following a chain-structured workflow. During each agent interaction between subtasks, an uncertainty quantification module monitors the generated response. If the uncertainty score u exceeds the phase-specific threshold $\tau_\phi$ (Yes branch), web retrieval is triggered to augment the available context before the agent proceeds with the response. Otherwise, when $u \leq \tau_\phi$ (No branch), the interaction continues through the standard workflow and advances to the next subtask.}
    \label{fig:ua_chatdev_architecture}
\end{figure*}

Software development is a complex, multi-phase engineering discipline encompassing requirements analysis, architectural design, implementation, code review, testing, and maintenance~\cite{wang2024llm,hou2024llm4se}. Traditional software development processes rely heavily on human expertise throughout each stage, making them time-consuming and susceptible to bottlenecks as system complexity continues to increase~\cite{hou2024llm4se,zhang2025unseen}. Test-debug cycles alone can account for up to 50\% of project costs~\cite{boehm2005software}, while the semantic gap between natural-language requirements and correct, executable code requires reasoning over ambiguity, context, and domain knowledge-capabilities that conventional rule-based tools often struggle to provide~\cite{lian2024imperfect,wang2025codegenErrors}. Recent advances in large language models (LLMs) have demonstrated significant potential for automating various software development tasks~\cite{wang2024llm,hou2024llm4se}, thereby alleviating these challenges and improving development efficiency~\cite{du2024classeval,zhang2025unseen}.

LLMs have demonstrated strong capabilities in role-playing, following multi-turn instructions, and generating programs from natural-language descriptions~\cite{hou2024llm4se, du2024classeval}, enabling autonomous multi-agent frameworks for software development through structured collaboration among specialized LLM agents~\cite{hou2024llm4se, he2025llm}. For example, MetaGPT~\cite{hong2024metagpt} assigns roles such as product manager, architect, engineer, and tester to automate the software development lifecycle, while ChatDev~\cite{qian2024chatdev} organizes LLM-based agents (e.g., CEO, CTO, programmer, reviewer, and tester) into a virtual software company for collaborative development. Qiu \textit{et al.}~\cite{qiu2025locobench} introduces a benchmark for assessing LLM agents on long-context software engineering workflows involving large codebases and multi-step interactions, and Rontogiannis \textit{et al.}~\cite{rontogiannis2026interactive} evaluates interactive LLM agents on multi-requirement software engineering tasks.

Although LLM-based multi-agent systems have made significant progress in automating software development, reducing development costs, and mitigating semantic gaps through collaboration among multiple specialized agents, current frameworks often assume that each agent’s output is reliable without explicitly modeling uncertainty~\cite{hong2024metagpt, qian2024chatdev, qiu2025locobench, rontogiannis2026interactive}. This limitation can lead to hallucination propagation, where incorrect requirements interpretation, architectural decisions, code generation, API usage, or test cases produced by one agent may be accepted and propagated by downstream agents. For example, a requirement analysis agent may misunderstand user needs, a coding agent may generate non-existent libraries or insecure implementations, and a testing agent may fail to detect hidden defects due to insufficient context. Uncertainty measurement enables agents to identify low-confidence decisions, trigger additional verification, request feedback from other agents, and allocate computational resources to high-risk tasks, thereby improving the reliability and efficiency of collaborative software development.

This study proposes  UA-ChatDev, an uncertainty-aware multi-agent software development framework that integrates uncertainty quantification into agent interactions to improve the reliability of LLM-based software generation. It introduces a lightweight uncertainty module that estimates response confidence using token-level log probabilities and detects unreliable intermediate outputs before they propagate across development phases. A task-specific threshold calibration strategy enables adaptive retrieval augmentation for uncertain responses, reducing hallucination propagation while preserving the efficiency of collaborative agent workflows. We conduct comprehensive experiments on the SRDD benchmark to evaluate the effectiveness of uncertainty quantification in multi-agent software development. The results demonstrate that the proposed framework consistently outperforms existing agent-based approaches across completeness, executability, consistency, and overall quality metrics, with particularly significant improvements in code execution reliability. Ablation studies and communication analyses further validate that uncertainty-aware interactions enhance agent coordination, reduce unreliable intermediate outputs, and improve the robustness of LLM-based software generation.

Our contributions are summarized below:

\begin{itemize}
\item  We propose UA-ChatDev, an uncertainty-aware extension of the ChatDev multi-agent software development framework that integrates uncertainty quantification into agent-to-agent interactions. Unlike existing multi-agent LLM systems for software development that assume generated responses are equally reliable, UA-ChatDev continuously evaluates the confidence of intermediate agent outputs during design, coding, and testing phases. 

\item We introduce a lightweight uncertainty quantification mechanism based on token-level log probabilities and task-specific threshold calibration to the software development automation. The uncertainty module estimates the reliability of each agent response, compares it with a calibrated phase-aware threshold, and selectively triggers external knowledge retrieval when uncertainty exceeds the acceptable level. 

\item We conduct comprehensive experiments on the SRDD benchmark to evaluate the effectiveness of uncertainty quantification in multi-agent software development, demonstrating consistent improvements over existing agent-based approaches across completeness, executability, consistency, and overall quality. Ablation studies and communication analyses further verify that uncertainty-aware interactions enhance code execution reliability.

\end{itemize}

\section{Methodology}
\label{sec3:method}


We introduce UA-ChatDev, a reliable framework that extends the ChatDev pipeline~\cite{qian2024chatdev} by incorporating uncertainty quantification into agent interactions across subtasks. Unlike conventional approaches that treat LLM-generated responses with uniform confidence, UA-ChatDev introduces an uncertainty module that intercepts agent communications between subtasks, estimates the response uncertainty based on the model’s token-level log probabilities, compares the resulting uncertainty score against a calibrated phase-specific threshold, and triggers external knowledge retrieval when the response is considered unreliable. This section presents the overall workflow of the proposed framework, the uncertainty quantification mechanism, and the threshold calibration strategy.

\subsection{Workflow of Multi-Agent Software Development}

The proposed framework builds upon the ChatDev pipeline~\cite{qian2024chatdev} as shown in Figure~\ref{fig:ua_chatdev_architecture}, which organizes software development as a collaborative process among multiple specialized agents. Given an initial task requirement, the system decomposes the development process into a sequence of structured phases, including design, coding, and testing. Each phase consists of multiple subtasks that are completed through interactions among role-specific agents.

Given an initial user requirement $R$, the software development process is modeled as a sequence of development phases:

\begin{equation}
P = \{P_D,P_C,P_T\},
\end{equation}

where $P_D$, $P_C$, and $P_T$ denote the design, coding, and testing phases, respectively. Each phase consists of a set of role-playing agents  $A = \{a_1,a_2,\dots,a_n\}$ that collaborate through structured communication to generate and refine software artifacts.

For each phase $P_\phi$, agents communicate through a multi-turn dialogue chain. Given the input context $X_i$ of subtask $i$, the response generated by agent $a_i$ is formulated as:

\begin{equation}
m_i = f(a_i, X_i, M_{<i}),
\end{equation}

where $m_i$ denotes the response generated by agent $a_i$, and $M_{<i}$ represents the preceding conversation history. The complete communication sequence within a phase is represented as:

\begin{equation}
M_\phi = \{m_1,m_2,\dots,m_k\},
\end{equation}

which enables agents to exchange information and collaboratively complete the assigned subtask.

After completing a subtask, the generated response is transformed into an intermediate software:

\begin{equation}
r_{i} = g(m_i,X_i),
\end{equation}

where $r_{i}$ contains the decisions, generated contents, and task-specific information produced by subtask $i$. It enables downstream agents to continue the development process based on previous results.

In the design phase, the CEO agent and CTO agent collaborate to transform the initial requirement $R$ into a software design:

\begin{equation}
D = f_{\phi_D}(R, M_D; A_{\mathrm{CEO}},A_{\mathrm{CTO}}).
\end{equation}

The design  $D$ contains the architectural decisions and implementation specifications that guide subsequent development activities.

During the coding phase, the CTO agent coordinates with programmer agents to transform the design specification into executable code:

\begin{equation}
C = f_{\phi_C}(D, M_C; A_{\mathrm{CTO}},A_{\mathrm{Prog}}).
\end{equation}

This phase consists of multiple implementation subtasks, such as code generation and code completion. Programmer agents iteratively exchange responses and refine intermediate code artifacts.

After implementation, the testing phase evaluates the generated software:

\begin{equation}
T = f_{\phi_T}(C, M_T; A_{\mathrm{Review}},A_{\mathrm{Test}}).
\end{equation}

The reviewer agent inspects the generated code to identify potential defects, while the tester agent validates software functionality through test execution. The detected issues are transmitted back to programmer agents as feedback to allow programmer agents to revise and improve the implementation:

\begin{equation}
C^{*}=f_{\mathrm{repair}}(C,T;A_{\mathrm{Prog}}).
\end{equation}

Therefore, the overall software development workflow can be formulated as $R \rightarrow D \rightarrow C \rightarrow T \rightarrow C^{*}$, where each stage progressively transforms the requirement into a refined software solution through multi-agent collaboration and response transmission across subtasks. Throughout the entire workflow, agents communicate through a structured chat chain, where the output of one subtask serves as the input for the next subtask, enabling progressive refinement from requirements to a final software solution.

\subsection{Uncertainty Quantification}
\label{sec:uncertainty_methods}

Treating agent outputs as inherently reliable can potentially lead to hallucination propagation, where incorrect requirement interpretations, architectural decisions, code generation, API usage, or test cases produced by one agent are accepted and propagated by downstream agents. To address this issue, we introduce a module of uncertainty quantification between substasks based on  single-inference uncertainty estimation~\cite{han2024towards} to evaluate the reliability of intermediate software artifacts before they are propagated to downstream agents. This design reduces computational overhead while preserving the efficiency of multi-agent collaboration by avoiding unnecessary checks on every intermediate communication. By focusing on critical transition points, the framework can detect potential errors early and trigger targeted knowledge retrieval to improve the reliability of generated outputs.

Given a response $r_{i}$ comprising $n$ tokens with log-probabilities $\{p_1, \ldots, p_n\}$, we apply temperature-scaled softmax to obtain probability estimates for the token $k$ by

\begin{equation}
    z_k = \frac{\exp(p_k / t)}{\sum_j \exp(p_j / t)},
\end{equation}

where the temperature $t = 1.0$ by default. Afterwards, we employ the average below to quantify the response uncertainty.

\begin{equation}
    u = -\log \text{Avg}(z_1, \ldots, z_n).
\end{equation}

Specifically, for single-token responses such as yes/no planning decisions, no softmax is applied and uncertainty reduces to $u = |p_1|$, the absolute value of the top-1 token's log-probability.

%


An important practical distinction arises between proprietary and open-source LLM backends. Proprietary APIs such as OpenAI expose token log-probabilities directly in completion response. On contrary, open-source models such as Gemma 2 9B~\cite{team2024gemma} and Qwen2.5-Coder 7B~\cite{hui2024qwen2}, loaded locally via the HuggingFace Transformers library~\cite{wolf2020transformers}, do not expose log-probabilities by default. To obtain the required probability signal, we retain the full unnormalized score tensor for every generated token. Per-token log-probabilities are then derived by applying log-softmax over the vocabulary dimension:

\begin{equation}
    p_k = \log \frac{\exp(s_k)}{\sum_v \exp(s_v)}
\end{equation}

where $s_i$ is the raw logit for token $i$ and the sum runs over all vocabulary entries. For efficiency, only the top-$k$ log-probabilities per token position are retained and passed to the uncertainty estimator.

\subsection{Uncertainty Threshold}

A single global threshold is insufficient because various subtasks generate responses of widely varying length. For instance, code review comments span dozens of lines while design acknowledgements are often a single sentence. Applying a threshold for one subtask to another will systematically misfire, either suppressing retrieval where it is needed or triggering it where it is not.

Given a set of historical development log files in each subtask, we extract all uncertainty values from responses whose downstream code passed the subsequent subtask, labelled \textit{correct}. We sort these values and compute the $(1 - \alpha)$ quantile as the subtask threshold:

\begin{equation}
    \tau_\phi = Q_{1-\alpha}\bigl(\{u : \text{response in  the subtask}   \text{ is correct}\}\bigr)
\end{equation}

We use $\alpha = 0.10$, corresponding to the 90th percentile. When fewer than 50 correct responses are available for a subtask, the procedure falls back to a global default $\tau = 0.5$. 

%
%
%
%
\section{Experiments}
\label{sec4:experiments}

\subsection{Dataset}


We validate the proposed framework on the Software Requirement Description Dataset (SRDD)~\cite{qian2024chatdev}, which contains 1,200 natural language software tasks spanning five categories: Education, Work, Life, Game, and Creation. Each category includes 240 tasks, further organized into 40 subcategories with 30 tasks per subcategory. The dataset covers a wide range of scenarios, from simple utility programs to complex interactive applications, providing a comprehensive benchmark for evaluating software generation quality across diverse domains.

\subsection{Implementation and Baselines}


We employ two open-source LLMs, Gemma 2 9B~\cite{team2024gemma} and Qwen2.5-Coder 7B~\cite{hui2024qwen2}, to implement our proposed framework, with the temperature parameter set to 0.2. The framework consists of five development subtasks spanning three phases (Design, Coding, and Testing), allows a maximum of ten communication rounds per subtask, and enables communicative dehallucination during Code Completion, Code Review, and Testing. 

In addition, we select three state-of-the-art (SOTA) LLM-based software development agents, GPT-Engineer\cite{osika}, MetaGPT\cite{hong2023metagpt}, and ChatDev~\cite{qian2024chatdev}, as baseline systems for performance comparison. GPT-Engineer represents a single-agent software generation paradigm, where an LLM autonomously transforms user requirements into executable code through iterative prompting. MetaGPT extends this paradigm by introducing a multi-agent framework with predefined software engineering roles, such as product manager, architect, engineer, and tester, enabling structured collaboration throughout the development lifecycle. ChatDev further advances multi-agent software development by organizing agents into a simulated software company and facilitating communication among specialized roles for requirements analysis, design, implementation, and testing. These baselines cover representative single-agent and multi-agent approaches, allowing us to evaluate whether uncertainty-aware agent collaboration can improve the reliability and quality of LLM-driven software generation.


\subsection{Evaluation Metrics}

We employ four metrics~\cite{qian2024chatdev} to comprehensively evaluate the software generation performance. \textbf{Completeness} ($C$) is a binary metric that indicates whether the generated code is fully implemented without any unimplemented placeholders or incomplete components. \textbf{Executability} ($E$) is a binary metric that measures whether the generated code can successfully run to completion without runtime errors. \textbf{Consistency} ($K$) evaluates the semantic alignment between the task requirement and the generated implementation by measuring the cosine similarity between their CodeBERT~\cite{feng2020codebert} embeddings. Finally, Overall \textbf{Quality} ($Q$) is defined as the product of Completeness, Executability, and Consistency, requiring the generated software to simultaneously satisfy implementation completeness, successful execution, and semantic correctness.

\subsection{Results and Discussions}

\subsubsection{Overall Performance Analysis}

\begin{table}[!ht]
\centering
\caption{Performance comparison of the proposed framework with existing agent-based software development systems on the SRDD benchmark. \faUser~= single-agent; \faUserFriends~= multi-agent. }
\label{tab:overall}
\begin{tabular}{lccccc}
\toprule
\textbf{System} & \textbf{Paradigm} & \textbf{C} & \textbf{E} 
& \textbf{K} & \textbf{Q} \\
\midrule
GPT-Engineer~\cite{osika} & \faUser   & 0.502 & 0.358 & 0.789 & 0.142 \\
MetaGPT~\cite{hong2023metagpt} & \faUserFriends     & 0.483 & 0.415 & 0.760 & 0.152 \\
ChatDev~\cite{qian2024chatdev} & \faUserFriends     & 0.560 & 0.880 & 0.802 & 0.395 \\
\midrule
Ours (Gemma 2) & \faUserFriends & 0.767 & 0.840 & 0.929 & 0.596 \\
Ours (Qwen 2.5) & \faUserFriends & \textbf{0.818} & \textbf{0.859} & \textbf{0.927} & \textbf{0.649} \\
\bottomrule
\end{tabular}
\end{table}

As shown in Table~\ref{tab:overall}, the results demonstrate that UA-ChatDev consistently outperforms existing agent-based software development systems on the SRDD benchmark across all evaluation metrics. Compared with SOTA models including GPT-Engineer, MetaGPT, and ChatDev, UA-ChatDev (ours) achieves the highest Overall Quality (Q), reaching 0.649 with Qwen 2.5 and 0.596 with Gemma 2, compared with 0.395 for ChatDev. This improvement indicates that incorporating the uncertainty quantification mechanism into agent collaboration substantially enhances the reliability of LLM-based software generation.

Moreover, UA-ChatDev (ours) achieves significant improvements in other metrics, including Completeness (C), Executability (E), and Consistency (K), highlighting the effectiveness of uncertainty-guided interaction. For instance, compared with ChatDev, UA-ChatDev improves Completeness from 0.560 to 0.818 and Consistency from 0.802 to 0.927. These improvements suggest that uncertainty quantification helps detect unreliable intermediate outputs, enables adaptive refinement, and reduces the propagation of incomplete code across different software development phases.

Lastly, the consistent improvements across different LLM backbones with Gemma and Qwen demonstrate that UA-ChatDev (ours) provides a model-agnostic reliability enhancement for multi-agent software development. Unlike existing multi-agent frameworks that often assume all agent-generated responses are equally trustworthy, UA-ChatDev (ours) explicitly models response confidence through uncertainty quantification and performs targeted intervention when uncertainty is high. This uncertainty-aware multi-agent framework improves software generation quality, as well as preserves the advantages of agent-based development workflows.

\begin{table}[!ht]
\centering
\caption{Comparison of software generation efficiency and complexity across different agent-based software development frameworks. \#Tokens (number of tokens used), \#Files (number of code files generated), and \#Lines (total lines of code across all files) in the software generation process.}
\label{tab:software_stats}
\begin{tabular}{lrrrr}
\toprule
\textbf{Method} & \textbf{Duration} (s) & \textbf{\#Tokens} & \textbf{\#Files} & \textbf{\#Lines} \\
\midrule
GPT-Engineer & 15.6000 & 7182.5333 & 3.9475 & 70.2041 \\
MetaGPT & 154.0000 & 29278.6510 & 4.4233 & 153.3000 \\
ChatDev & 148.2148 & 22949.4450 & 4.3900 & 144.3450 \\
\midrule
Ours (Gemma 2) & 279.9430 & 35725.2688 & 4.0327 & 118.0042 \\
Ours (Qwen 2.5) & 305.5272 & 40490.1821 & 5.6433 & 168.0769 \\
\bottomrule
\end{tabular}
\label{tab:software_statistics}
\end{table}

Additionally, results in Table~\ref{tab:software_stats} indicates that UA-ChatDev introduces additional refinement and verification steps, which may increase generation complexity depending on the underlying LLM backbone. While UA-ChatDev with Qwen 2.5 produces more extensive software artifacts than ChatDev, the Gemma 2 variant generates fewer files and lines of code, suggesting that uncertainty-aware quantification primarily improves reliability through quality-oriented refinement rather than simply increasing output size. Nevertheless, the additional uncertainty quantification and adaptive intervention mechanisms introduce computational overhead in terms of generation time (\textbf{Duration}) and token consumption (\textbf{\#Tokens}).

\subsubsection{Ablation Study}

\begin{table}[!ht]
\centering
\caption{Ablation study on the impact of uncertainty quantification. Full model refers to UA-ChatDev with the uncertainty quantification mechanism, while w/o Uncertainty indicates the variant without uncertainty quantification.}
\label{tab:uncertainty}
\begin{tabular}{llcccc}
\toprule
\textbf{LLMs} & \textbf{Condition} & \textbf{C} & \textbf{E} 
& \textbf{K} & \textbf{Q} \\
\midrule
\multirow{2}{*}{Gemma 2} 
& w/o Uncertainty & 0.758 & 0.794 & 0.914 & 0.557 \\
& Full Model  & \textbf{0.767} & \textbf{0.840} & \textbf{0.929} & \textbf{0.596} \\
\midrule
\multirow{2}{*}{Qwen 2.5} 
& w/o Uncertainty  & 0.807 & 0.779 & 0.926 & 0.578 \\
& Full Model  & \textbf{0.818} & \textbf{0.859} & \textbf{0.927} & \textbf{0.649} \\
\bottomrule
\end{tabular}
\end{table}

This section examines the impact of uncertainty quantification. Results in Table~\ref{tab:uncertainty} demonstrates that uncertainty quantification plays a critical role in improving the reliability of multi-agent software development. Compared with the variant without uncertainty quantification, Full Model consistently achieves better performance across all evaluation metrics (C, E, K, and Q) for both backbone LLMs, including Gemma 2 and Qwen 2.5. This indicates that incorporating uncertainty estimation into agent interactions effectively improve the overall quality of generated software artifacts.

Additionally, the improvements are particularly significant in execution (E) and quality (Q) metrics, suggesting that uncertainty-aware communication enables agents to make more reliable decisions before passing information to downstream phases. For example, Full Model with Qwen 2.5 improves from 0.779 to 0.859 in execution and from 0.578 to 0.649 in quality after introducing uncertainty quantification. These results confirm that uncertainty quantification enables more robust reasoning, validation, and refinement of intermediate solutions.

\subsubsection{Communication Analysis}

\begin{figure}[!ht]
\centering
  \includegraphics[width=0.4\textwidth]{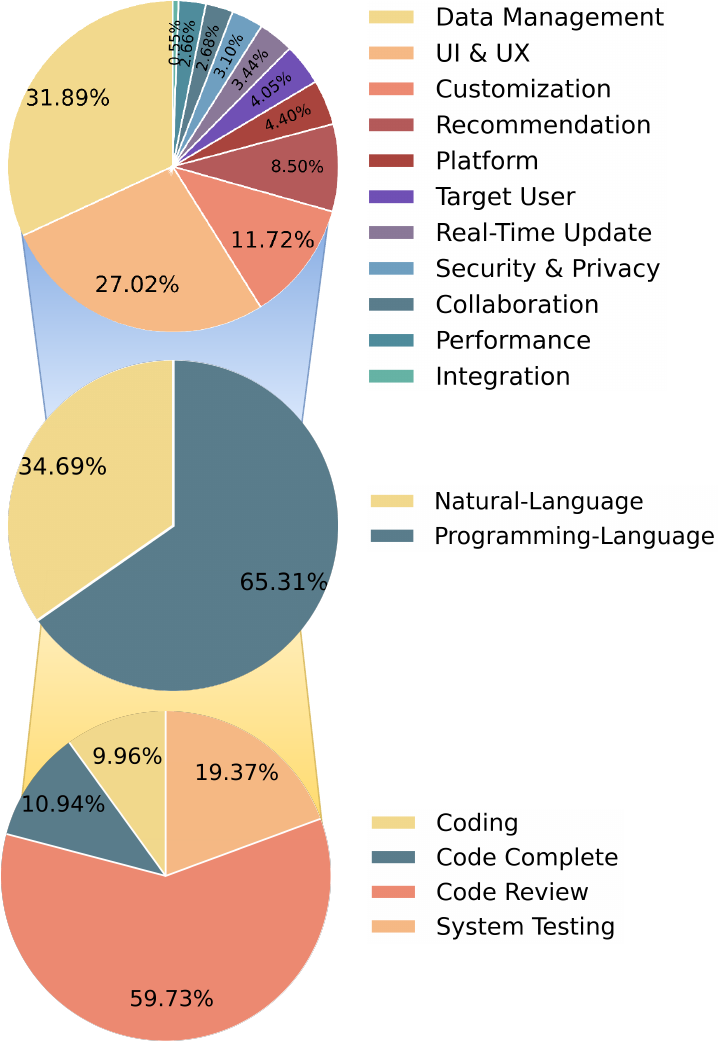} 
  \caption{Utterance distribution of agent communications throughout the entire development process.}
  \label{fig:untterance}
\end{figure}

\begin{figure*}[!ht]
\centering
  \includegraphics[width=1\textwidth]{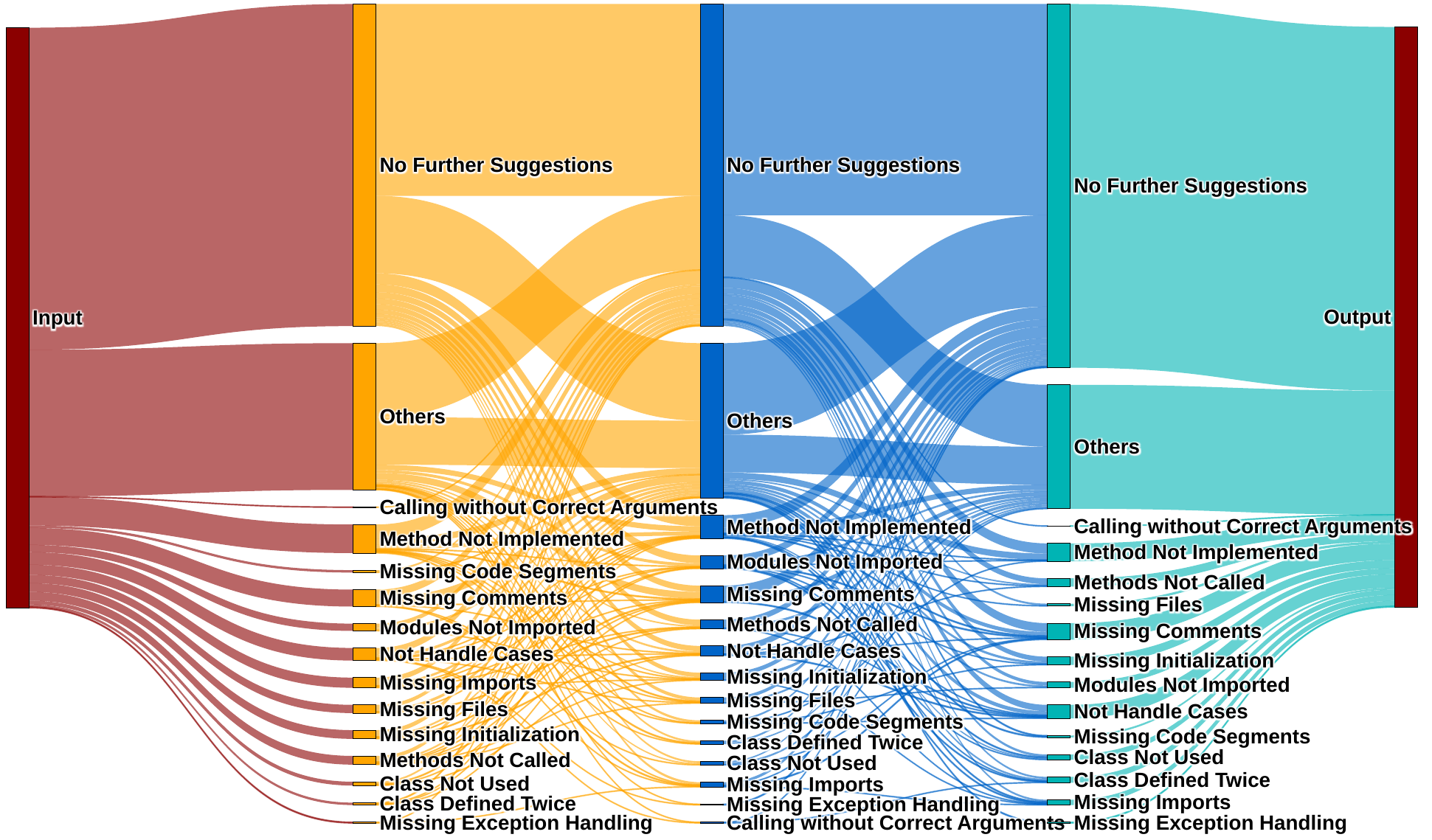} \\
  \caption{Illustration of the distribution of suggestions made by a reviewer agent during a multi-round reviewing process, where each sector in the chart represents a different category of suggestion.}
  \label{fig:suggestion}
\end{figure*}

\begin{figure*}[!ht]
\centering
  \includegraphics[width=1\textwidth]{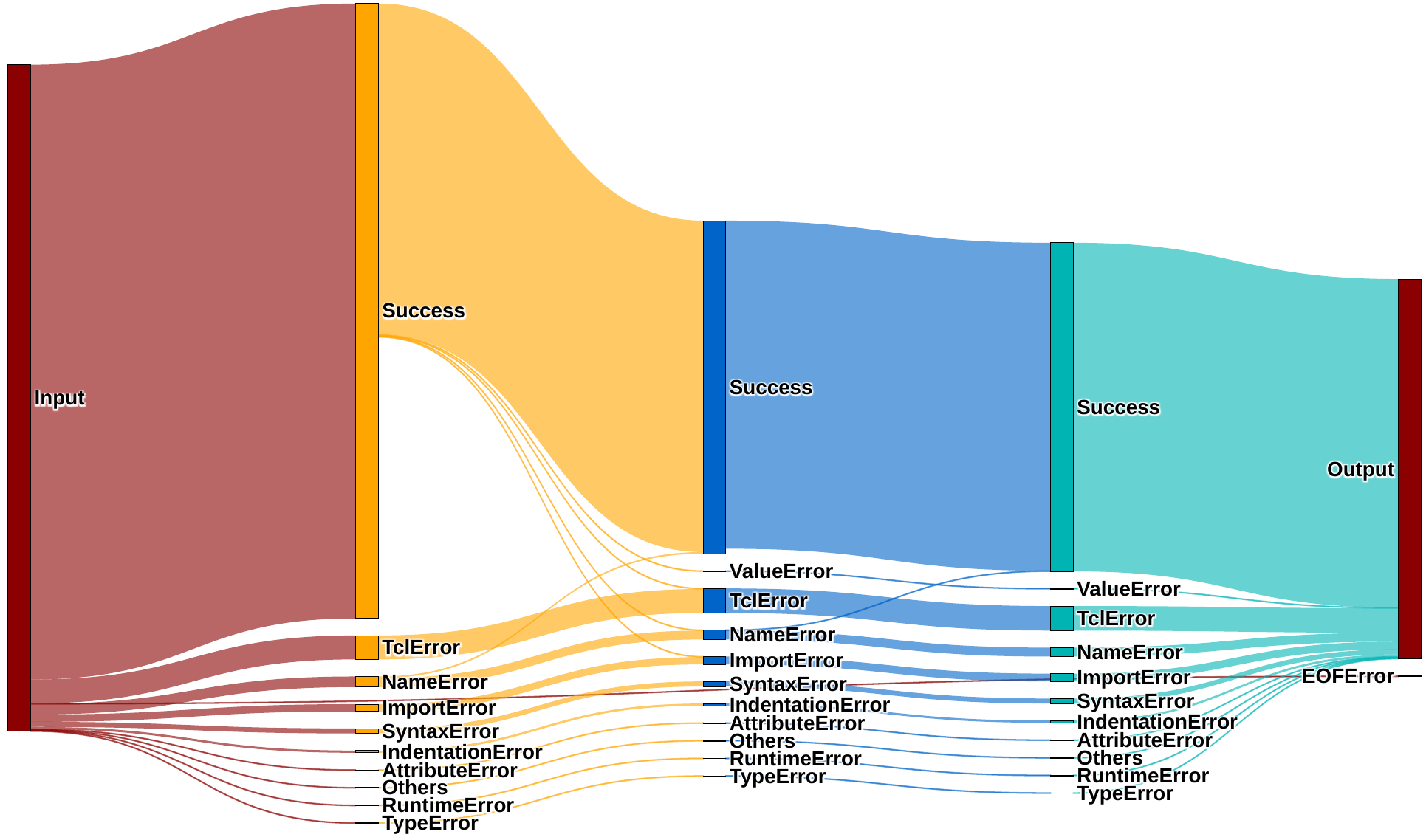} \\
  \caption{Diagram of the progression of iterations in a multi-round testing process, where each colored
column represents a dialogue round, showcasing the evolution of the solution through successive stages of testing.}
  \label{fig:test_error}
\end{figure*}

This section aims to analyze the content of agent communications to understand linguistic effects. First, Figure~\ref{fig:untterance} presents the utterance distribution across the development process.  Overall development (top plot: communication categories) shows that agent communication covers a wide range of software engineering activities, with the largest portions involving Data Management (31.89\%), UI \& UX (27.02\%), and Customization (11.72\%). This indicates that multi-agent software development requires extensive coordination not only for implementation but also for understanding requirements, organizing data, and adapting system functionalities. Although categories such as security, integration, and performance account for smaller proportions, they may still introduce critical risks because mistakes in these stages can negatively impact the final software quality.

The programming language distribution (middle plot) demonstrates that agent communication is primarily focused on Programming Language-related interactions (65.31\%), while Natural Language communication contributes 34.69\%. This suggests that agents frequently exchange technical information related to code generation, implementation, and debugging. At the same time, the significant proportion of natural language interactions reflects the challenge of translating ambiguous user requirements into precise software specifications. 

The coding workflow distribution (bottom plot) further reveals that Code Review dominates the coding-related interactions (59.73\%), followed by Coding (19.37\%), Code Complete (10.94\%), and System Testing (9.96\%). The high proportion of code review activities suggests that agent collaboration involves continuous validation and refinement of generated artifacts rather than simple code generation. This observation supports the role of uncertainty quantification in improving software reliability by helping agents detect low-confidence outputs, prioritize verification efforts, and reduce error propagation throughout the development lifecycle.

Second, Figure~\ref{fig:suggestion} demonstrates the effectiveness of uncertainty-aware reviewer agents in improving the reliability of multi-round software review. The reviewer agents identify various implementation-level issues, such as \textbf{Missing Imports}, \textbf{Missing Files}, and \textbf{Not Handel Cases} which directly affect whether generated programs can successfully execute, which is aligned with the observations in Table~\ref{tab:uncertainty}. By incorporating uncertainty estimation into agent interactions, the system can recognize low-confidence code generation decisions and trigger additional verification or refinement before producing the final artifact.

The persistence of execution-related defect categories across review rounds indicates that these issues require iterative validation rather than simple detection. For example, \textbf{Calling without Correct Arguments} may not always be resolved in a single step because it requires understanding the relationship between generated functions, interfaces, and program logic. The uncertainty-aware mechanism helps agents allocate additional reasoning and verification effort to ambiguous code components, reducing the likelihood that uncertain implementations are propagated into the final executable program.

Third, Figure~\ref{fig:test_error} illustrates the evolution of generated code execution outcomes across multiple testing rounds via the uncertainty-aware multi-agent system. The error transitions reveal that uncertainty quantification helps agents address different sources of execution failures, including \textbf{SyntaxError}, \textbf{ImportError}, \textbf{TypeError}, and \textbf{ValueError}. These failures often arise from incorrect program structures, missing dependencies, or inconsistent function interactions. Instead of accepting generated code based only on surface-level correctness, uncertainty-aware agents use execution feedback to locate potentially unreliable components and iteratively revise them, improving the probability of successful execution.

\section{Related Work}
\label{sec2:background}

Multi-agent frameworks have emerged as a compelling approach to automating software development by distributing the cognitive load of complex engineering tasks across specialised LLM instances. For instance, GPT-Engineer~\cite{osika} provides a simpler, single-agent alternative demonstrating the feasibility of one-step repository generation. MetaGPT~\cite{hong2023metagpt} assigns fixed roles with standardized operating procedures, achieving high structural coverage but lacking adaptive dialogue.  GameGPT~\cite{chen2023gamegpt} focuses on the multi-agent approach for game development. ChatDev~\cite{qian2024chatdev} introduced the chat-chain paradigm for role-playing LLM agents in software development, the framework our work extends. 


Beyond commercial LLMs, open-source code-centric LLMs have emerged as effective foundations for automated software engineering tasks. For example, CodeGen\cite{nijkamp2022codegen} explores multi-turn program synthesis, showing that iterative interaction improves the quality of generated programs. CodeLlama\cite{roziere2023code} adapts LLaMA 2 for software engineering through large-scale code pre-training and infilling-based objectives, supporting diverse code completion and generation scenarios. CodeGemma~\cite{team2024codegemma} enhances the Gemma family with code-specific training strategies and instruction tuning, achieving strong results on widely adopted benchmarks including HumanEval and MBPP.


Meanwhile, recent studies have explored uncertainty estimation and uncertainty-aware mechanisms to improve the reliability of LLMs.  For example, Malinin and Gales~\cite{malinin2020uncertainty} establish entropy-based uncertainty estimation for autoregressive models. Kuhn \textit{et al.}~\cite{kuhn2023semantic} introduce semantic entropy to account for paraphrase equivalence in free-form answers. Yang \textit{et al.}~\cite{yang2023improving} apply uncertainty-aware in-context learning to improve reliability. FLARE~\cite{jiang2023active} uses token probabilities to trigger retrieval mid-generation. \uala{}~\cite{han2024towards}  synthesizes these into a complete agent framework.

Although these approaches have made significant progress toward automating software development through LLM-based agent systems, they have not fully explored the role of uncertainty quantification in agent interactions and how unreliable intermediate outputs may propagate across different development phases. Existing frameworks typically assume that agent-generated responses are equally reliable and focus primarily on improving collaboration strategies, task decomposition, and role specialization. 

Moreover, recent studies on retrieval-augmented generation (RAG) have been applied to code generation primarily for API documentation retrieval~\cite{patil2024gorilla} and repository-level context augmentation~\cite{chen2021evaluating}. In contrast, our uncertainty-triggered retrieval mechanism can be viewed as a form of selective RAG, where rather than retrieving external knowledge for every generation request, it activates retrieval only when the model’s uncertainty indicates insufficient internal knowledge. This strategy reduces unnecessary retrieval overhead while preserving quality improvements for scenarios where additional information is most beneficial.


%
\section{Conclusion and Future Work}
\label{sec6:conclusion}
This paper presents UA-ChatDev, an uncertainty-aware multi-agent software development framework that integrates uncertainty quantification into agent interactions to improve the reliability of LLM-generated software. By estimating the confidence of intermediate agent outputs and applying phase-aware uncertainty calibration, UA-ChatDev enables adaptive verification and reduces the propagation of unreliable decisions throughout the software development lifecycle. Extensive experiments on the SRDD benchmark demonstrate that UA-ChatDev consistently outperforms existing agent-based software development approaches, with notable improvements in code execution reliability, completeness, consistency, and overall quality. These findings highlight the importance of uncertainty-aware collaboration for building robust, trustworthy, and scalable LLM-driven software engineering systems.

In the future, we plan to explore selective retrieval strategies that fetch concise summaries rather than full document snippets, which potentially reduces the token overhead of triggered interventions without sacrificing the quality benefit. Second, per-phase uncertainty budgeting would allow the framework to concentrate retrieval in the phases, such as Code Review and Testing, where it yields the greatest return, rather than applying a uniform policy across the pipeline. Third, extending UN-ChatDev to multi-model agent teams, where different roles are served by different LLMs, would allow specialized models to be paired with targeted uncertainty thresholds calibrated to their individual generation characteristics. 
%

\section*{Acknowledgment}
This research work is partially supported by the U.S. NSF under award number 2235731, 2323419, 24018601 and by the Army Research Office (ARO) under cooperative agreement W911NF-24-2-0133. The views and conclusions contained in this document are those of the authors and should not be interpreted as representing the official policies, either expressed or implied, of NSF or ARO or the U.S. Government. The U.S. Government is authorized to reproduce and distribute reprints for Government purposes notwithstanding any copyright notation herein. Additionally, the authors acknowledge the use of AI-based tools, such as ChatGPT, for assistance in editing, grammar enhancement, and spelling checks during the preparation of this manuscript.



\bibliographystyle{IEEEtran}
\bibliography{References}
%



\end{document}